\documentclass[letterpaper, 10 pt, journal, twoside]{IEEEtran}

\usepackage{amsmath,amsfonts}
\usepackage{algorithm}
\usepackage{array}
\usepackage[caption=false,font=normalsize,labelfont=sf,textfont=sf]{subfig}
\usepackage{textcomp}
\usepackage{stfloats}
\usepackage{url}
\usepackage{verbatim}
\usepackage{graphicx}
\usepackage{cite}
\usepackage{amssymb, physics, amsthm, bbm}
\usepackage[bookmarks=true]{hyperref}
\usepackage[nameinlink]{cleveref}
\Crefname{figure}{Fig.}{Figs.}
\usepackage{booktabs}
\usepackage{fontawesome}
\usepackage{algpseudocode}

\theoremstyle{definition}
\newtheorem{problem}{Problem}
\newcommand{\controller}{\textit{c}}
\newcommand{\symb}{\bot}

\usepackage{fancyhdr}
\fancypagestyle{preprintheader}{
  \fancyhf{}
  
  \fancyhead[C]{%
    \vspace{-0.0cm}%
    \parbox{\textwidth}{%
      \centering \scriptsize 
      \copyright~2026 IEEE. Personal use of this material is permitted. Permission from IEEE must be obtained for all other uses, in any current or future media, including reprinting/republishing this material for advertising or promotional purposes, creating new collective works, for resale or redistribution to servers or lists, or reuse of any copyrighted component of this work in other works. RA-L preprint version. DOI: \href{https://doi.org/10.1109/LRA.2026.3685463}{10.1109/LRA.2026.3685463}
    }%
  }
}

\begin{document}

\title{Cross-Entropy Optimization of Physically\\ Grounded Task and Motion Plans}

\author{Andreu Matoses Gimenez$^{1}$, Nils Wilde$^{2}$, Chris Pek$^{1}$, and Javier Alonso-Mora$^{1}$
\thanks{Manuscript received: November 2, 2025; Revised February, 22, 2026; Accepted March, 23, 2026.}
\thanks{
This work was supported by the European Union through ERC INTERACT, Grant 101041863. Views and opinions expressed are, however, those of the author(s) only and do not necessarily reflect those of the European Union. Neither the European Union nor the granting authority can be held responsible for them.}
\thanks{$^{1}$The authors are with the Department of Cognitive Robotics, Delft University of Technology, The Netherlands
        {\tt\footnotesize  \{a.matosesgimenez, c.pek, j.alonsomora\}@tudelft.nl}}%
\thanks{$^{2}$Nils Wilde is with the Faculty of Computer Science, Dalhousie University, Halifax, Canada
        {\tt\footnotesize  nils.wilde@dal.ca}}%
}


\markboth{IEEE Robotics and Automation Letters. Preprint Version. Accepted March, 2026}
{Matoses Gimenez \MakeLowercase{\textit{et al.}}: Cross-Entropy Optimization of Physically Grounded Task and Motion Plans}


\maketitle
\thispagestyle{preprintheader}

\begin{abstract}
Autonomously performing tasks often requires robots to plan high-level discrete actions and continuous low-level motions to realize them. 
Previous TAMP algorithms have focused mainly on computational performance, completeness, or optimality by making the problem tractable through simplifications and abstractions.
However, this comes at the cost of the resulting plans potentially failing to account for the dynamics or complex contacts necessary to reliably perform the task when object manipulation is required.
Additionally, approaches that ignore effects of the low-level controllers may not obtain optimal or feasible plan realizations for the real system. 
We investigate the use of a GPU-parallelized physics simulator to compute realizations of plans with motion controllers, explicitly accounting for dynamics, and considering contacts with the environment. 
Using cross-entropy optimization, we sample the parameters of the controllers, or actions, to obtain low-cost solutions. 
Since our approach uses the same controllers as the real system, the robot can directly execute the computed plans. We demonstrate our approach for a set of tasks where the robot is able to exploit the environment's geometry to move an object. Website and code: 
\href{https://andreumatoses.github.io/research/parallel-realization}{\nolinkurl{andreumatoses.github.io/research/parallel-realization} \faGithub}

\end{abstract}


\section{Introduction}
\IEEEPARstart{J}{ointly} planning tasks and motions for real-world robots presents significant challenges. It involves determining both what to do, i.e., the sequence of high-level actions, and how to do it, i.e., the specific motions associated with each action.
This joint problem is usually complex as early actions can impact the performance of later actions or even the overall plan feasibility. 
Task and Motion Planning (TAMP) methods aim to simultaneously satisfy symbolic constraints (relations between discrete, high-level actions in the task plan) and geometric constraints (requirements on their continuous, physical realization)~\cite{toussaint_logic-geometric_2015,garrett_integrated_2020}.

\begin{figure}[t]
    \centering
    \includegraphics[width=0.8\linewidth]{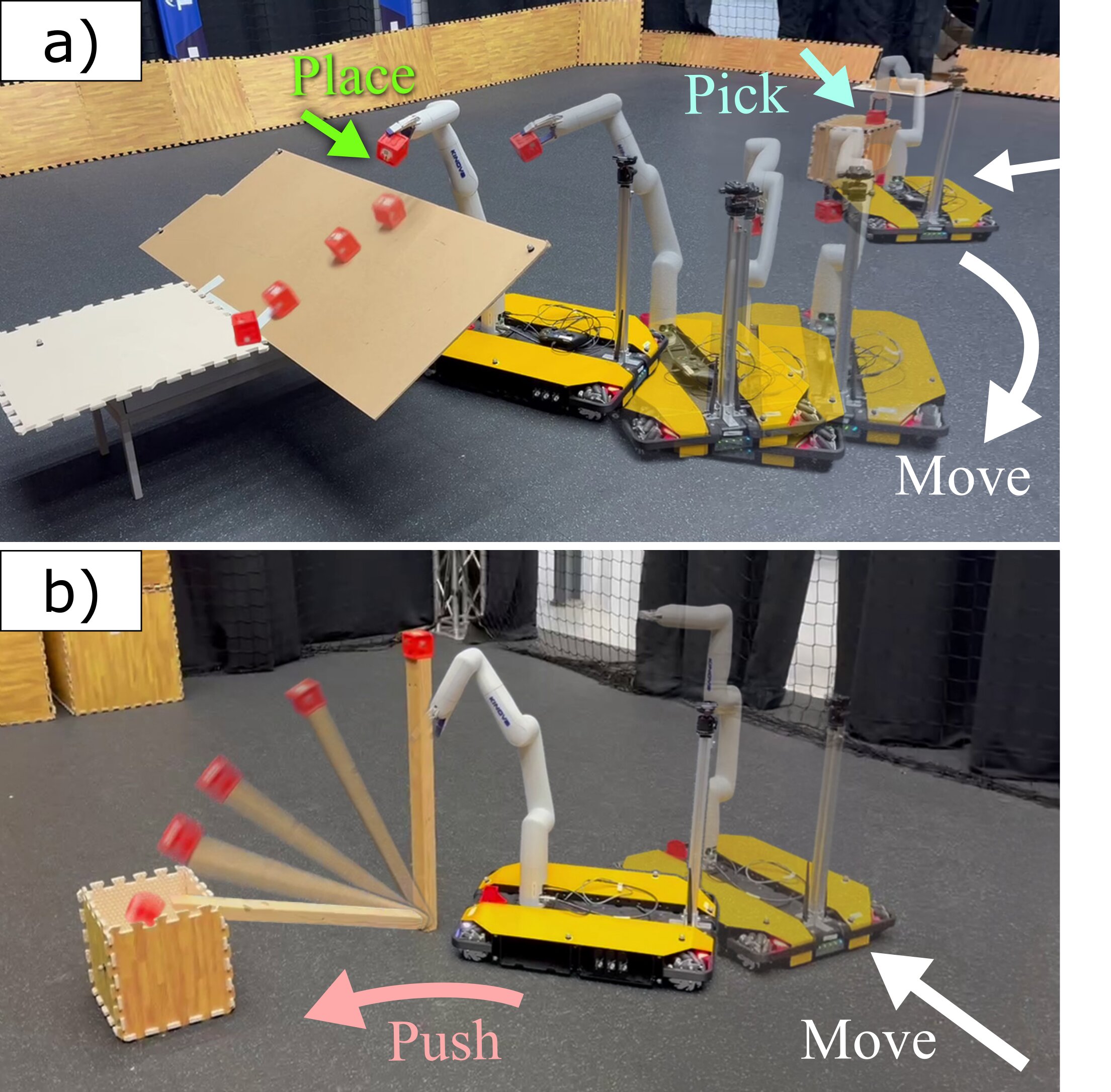}
    \caption{(a) Mobile manipulator performing a sequential pick and place task. The robot can exploit the obstacle's slope to find a faster solution, instead of going around. (b) Performing a move and push to put the block in the box.}
    \label{fig:intro}
\end{figure}

Consider, for example, a mobile manipulator that needs to move an object from $\mathit{Table}_1$ to $\mathit{Table}_2$ and finally move to a dedicated $\mathit{Exit}$ location, as shown in \Cref{fig:intro}a. 
In such a scenario, the robot needs to move in sequence to $\mathit{Table}_1$, perform a grasping action, move to $\mathit{Table}_2$ while holding the object, perform a placing action, and finally move to the $\mathit{Exit}$. 
To perform this autonomously, the robot would need to a) compute a high level plan, given by a \textit{sequence of actions}, while considering its capabilities and its goal, and b) decide how to perform each action, by selecting a suitable \textit{low-level controller}. The robot could, for example, approach $\mathit{Table}_1$ from the right or from the left side, it could go around the obstacle in many different ways, or even better, it could exploit the geometry of the ramp-shaped obstacle to achieve its goal faster. 
However, the robot's ability to decide how to perform actions is limited by the type of controllers available and the \textit{action parameters} that define the motions generated by the controllers. For example, grasp actions can use full-body geometric controllers which take as input desired final end-effector poses. Move actions can use path planning algorithms such as $A^*$ with a final position as input parameter. 

Solving these TAMP problems is computationally expensive and could require extensive modeling. 
The robot has to come up with a sequence of actions for the plan, suitable action parameters, and also evaluate whether such a plan is feasible and efficient.
Consequently, TAMP problems require the robot to consider the large combinatorial set of possible plans that accomplish the high-level task specification, as well as the set of possible action parameters. 
To deal with such large planning spaces in a computationally efficient way, most TAMP algorithms use simplifications and abstractions, such as sampling-based discretization of the state space \cite{garrett_pddlstream_2020}, simplified dynamics \cite{thomason_task_2022}, or limited modeling of contacts \cite{toussaint_differentiable_2018}, if any.
Furthermore, sampling-based TAMP approaches typically rely on rejection-sampling ``streams'' (generators) to synthesize continuous parameters. While effective for geometric constraints, these streams become inefficient when valid parameters lie in narrow manifolds determined by complex dynamics.

These assumptions and simplifications come at the cost that the resulting plans may not adequately account for the dynamics or complex contacts necessary to reliably perform the task. 
Additionally, when ignoring the effect of the robot's low-level controllers, such approaches may obtain suboptimal or even infeasible plans for the real system.

We investigate the use of parallelized physics simulators \cite{makoviychuk2021isaac} to compute low-cost realizations of a plan with realistic controllers.
The combination of such simulations with efficient and parallelizable controllers allows us to consider the dynamics, complex contacts, and limitations of the controllers themselves when evaluating plan realizations~\cite{noseworthy2024forge,xu2022accelerated}. We can see in \Cref{fig:intro}a how the robot takes advantage of the ramp to let the block slide to the table, without the need to go around, and \Cref{fig:intro}b, where the robot is able to push a stick to let the block fall into a box. 

To this end, we formulate the task planning problem using the widely used Planning Domain Definition Language (PDDL) \cite{mcdermott_pddl-planning_1998} due to its expressivity and the availability of off-the-shelf symbolic planners.
We then associate appropriate controllers and their input parameters to the operators used in the PDDL formulation.
We explicitly frame the synthesis of these controller parameters as a ``Stream'' (or generative sampling strategy). However, instead of using standard rejection sampling, we implement this stream using Cross-Entropy optimization to actively search for feasible and low-cost parameters. This enables the discovery of plan realizations that exploit complex dynamics and contacts without significantly increasing the modeling effort at the symbolic level. Since our approach uses the same controllers as the real system, the robot can directly execute our realized plans.

\subsection{Related Work}

There are a variety of integrated TAMP solvers \cite{garrett_integrated_2020,zhao_survey_2024}, which can be categorized into constraint-based, sampling-based, or optimization-based TAMP solvers.

Constraint-based TAMP \cite{dantam_incremental_2018} uses general-purpose Satisfiability Modulo Theory (SMT) solvers, where motion constraint information is incrementally added at the task level. These methods focus on feasibility satisfaction of the resulting plan, and require detailed knowledge about the symbolic and geometric representations of the world, which need to be sufficient to ground a given plan.
Optimization-based TAMP solvers often divide the problem into hierarchical feasibility (and optimality) stages with progressively more detail to filter the space of symbolic plans. In particular, \cite{toussaint_logic-geometric_2015,toussaint_multi-bound_2017} use tree search over sequences of symbolic actions and evaluate the feasibility and optimality of each node with increasing detail as an heuristic. Detailed motion calculation is based on carefully crafted mathematical constraints and kinematic models that can be leveraged by specialized optimizers \cite{toussaint_newton_2014, holladay_force-and-motion_2019}. For some problems these solvers may fail to find a solution, and computational cost can increase prohibitively if more detail is considered, such as dynamics and contacts. 

Sampling-based TAMP methods focus on efficiently generating action parameters which can then be used at the task planning stage. In \cite{garrett_sampling-based_2018,garrett_pddlstream_2020}, conditional samplers are used in combination with Planning Domain Definition Language (PDDL) \cite{mcdermott_pddl-planning_1998} symbolic planners to incrementally generate PDDL sub-problems from discretized action parameters. \cite{thomason_task_2022} introduces an optimal sampling-based approach using asymmetric bidirectional sampling, which significantly improves the efficiency of finding action parameters for plans without the need for discretization. 
These methods require the creation of domain-specific samplers that generate constraint-satisfying configurations of the robot and objects to handle complex dynamic interactions. This requires considerable engineering effort when adapting to new problems without significant simplifications.
For instance, analytically modeling the physical constraints for a single, generic "place" action that covers all possible dynamic outcomes is often intractable. Consequently, prior methods typically require discretizing these interactions into finer-grained, explicitly modeled actions (e.g., requiring both \textit{place\_flat} and \textit{place\_sloped}).
Recent work also explores the use of sampling-based optimization to find plan realizations solving a non-linear program \cite{toussaint_nlp_2024,braun_stein_2024}. \cite{toussaint_nlp_2024} explores the use of specialized MCMC sampling strategies and how they could potentially be used in the context of TAMP. \cite{braun_stein_2024} investigates the use of evolutionary strategies to generate high-quality samples of these complex distributions. These non-linear sampling-based optimization approaches are particularly promising in TAMP settings where solutions lie in multi-modal, disconnected regions of the parameter space.

There is growing interest in integrating closed-loop controllers into TAMP solvers to facilitate real-world deployment.
In \cite{toussaint_sequence--constraints_2022} a sequence of actions is executed by switching between MPC controllers with the corresponding sets of constraints. Once the set of goal states are deemed unreachable, a backtracking mechanism brings the execution back to the last action from which the plan is feasible. 
Most recently, \cite{pan_task_2024} utilized pre-defined behaviors to handle informational uncertainty (e.g., occlusions). In contrast, we address dynamic complexity. By optimizing generic actions via physics simulation, our framework automatically discovers complex interactions (e.g., sliding) without requiring them to be explicitly encoded as separate symbolic behaviors.
In \cite{curtis_partially_2024}, a framework for uncertainty-aware TAMP is presented. The uncertainty in the execution of the controllers is estimated by simulating the actions and recording successes. These approaches can be implemented in real-world scenarios, as closed-loop controllers are expected to compensate for disturbances arising from unmodeled complex dynamics and contact interactions. However, while these bring TAMP solvers closer to practical deployments, planning while considering these effects remains a challenge.

\subsection{Contributions}

This paper contributes a task and motion planning method that accounts for the robot's dynamics and physical constraints, environment, contacts, and controllers. We

\begin{enumerate}
    \item
    combine PDDL~\cite{mcdermott_pddl-planning_1998} with information on low-level controllers and required parameters to formulate symbolic tasks plans, treating the parameter synthesis as a black-box stream;
    \item
    propose a cross-entropy optimization algorithm to optimize the parameters of the controllers that realize the candidate plan and minimize the cost function. 
   Cross-entropy optimization is performed in a simulator to consider dynamics, contacts, and the controllers' limitations when finding plan realizations; 
    \item
    demonstrate the method on both a pick-and-place and pushing task, in simulation and on a real robot. Our results show that we can find solutions that take advantage of contact with the environment, which would require explicit modeling in traditional methods.
\end{enumerate}

\section{Problem Formulation}

We consider TAMP problems, where a mobile manipulator needs to execute a sequence of $k$ actions, performed by low-level controllers, to achieve a high-level goal while minimizing a cost function. Each controller requires the input values of several parameters to fully define the resulting robot motion. 
Additionally, we focus on TAMP problems where the robot may be able to exploit the environment's geometry and dynamics, e.g., by making contacts, to find low-cost plans. 

The configuration space of the system is $\mathcal{X} = \mathcal{Q}  \times SE(3)^m$, where $ \mathcal{Q}$ denotes the configuration space of the robot, and there are $m$ objects in the environment.  
We use $\mathbf{x}_t \in \mathcal{X}$ to denote the state of the whole system at time $t$. 
We assume that the problem can be solved within a time horizon $T$, if a solution exists.  
The task specification $\psi_g$ encodes the goal, which is achieved if $\psi_g(\mathbf{x}_T) = \text{True}$ for the final time $T$. Formally, the problem is stated as follows.
\begin{figure*}[t]
    \centering  \includegraphics[width=0.75\linewidth,trim= 30 0 0 0, clip]{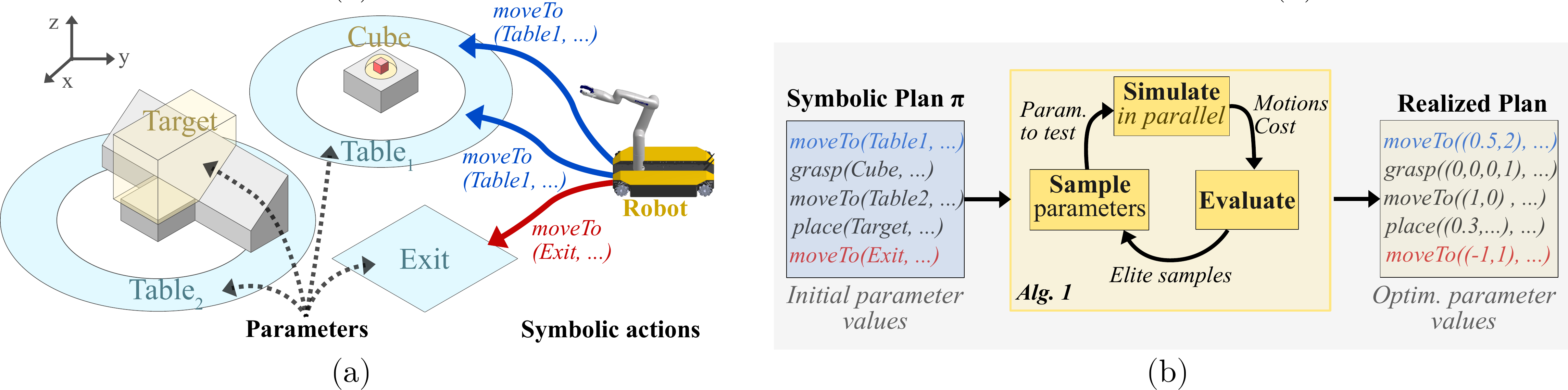}
    \caption{Overview of our TAMP framework. (a) The scenario specification involves defining symbolic actions the robot can perform, e.g., moving to something, and parameters for the actions, e.g., a region corresponding to Table1 or the Exit. Note that parameterized actions such as \emph{moveTo(Table1,\ldots)} may result in different motions, shown in blue. (b) We optimize the parameters of the symbolic plan using parallel simulations and cross entropy optimization.}
    \label{fig:realization_diagram}
\end{figure*}

\begin{problem}[Sequence of Controllers]
\label{prob:controllers_seq}
    Given a set of controllers $\Gamma$, cost functions \eqref{eq:optimization1_obj}, goal specification $\psi_g(\mathbf{x}_T)$, and system dynamics \eqref{eq:optimization1_dyn},
    we aim to find a sequence of $k$ controllers $\controller_{1:k}$, each selected from $\Gamma$, and their parameters $z_{1:k}$ that solve the following optimization problem:
\begin{subequations} \label{eq:optimization}
\begin{align}
\min_{\controller_{1:k}, z_{1:k}} \quad 
& \sum_{i=0}^{k} \sum_{t=t_{0,i}}^{T_i} J(x_t, \dot{x}_t, \controller_i, z_i) +J_{\text{end}}(x_T) 
\label{eq:optimization1_obj} \\[6pt]
\text{s.t.} \quad 
& x_{t+1} = f(x_t, \controller_i, z_i),
\label{eq:optimization1_dyn} \\[3pt]
& \psi_g(\mathbf{x}_T) = \text{True}
\label{eq:optimization1_goal}
\end{align}
\end{subequations}
where $t_{0,i}$ and $T_i$ are the initial and final time of controller $\controller_i$ execution. The next controller's execution will start thereafter, i.e., $t_{0,i+1} = T_i +1$. The number of controllers $k$ is not known a priori, and assumed to be bounded. 
\end{problem}

\section{Method}

Solving \Cref{prob:controllers_seq} requires simultaneously finding the discrete sequence of controllers $\controller_{1:k}$ and their continuous parameters $z_{1:k}$ in a high-dimensional planning space. 
Further, we must consider non-linear dynamics, as well as discontinuities introduced by contact events and controller switches.
These additional considerations make it intractable to use direct search methods to find the sequence of controllers that achieve the goal specification.

To address this issue, we propose to divide the problem into two levels, illustrated in~\Cref{fig:realization_diagram}. 
First, an abstracted symbolic task planning step, where we define the robot's actions and associated parameters for the task, and find a candidate sequence of actions and their associated controllers. 
This step effectively provides a good candidate for the assignment of $\controller_{1:k}$ in \Cref{prob:controllers_seq}, and narrows down the parameters.
We refer to this candidate plan as symbolic plan $\pi$.
Secondly, we perform cross-entropy optimization to determine values for the parameters that render the plan feasible with low cost.
We call this step finding a plan realization, which provides values to the parameters $z_{1:k}$ consistent with the task plan.
In this work, we assume that the candidate symbolic plan is physically feasible. We define high-level actions broadly (e.g., a generic \texttt{place} action), leaving the complex geometric details to be solved by the optimizer. This generality maximizes the chance of finding a working solution. While our optimization method could theoretically search over multiple candidate plans simultaneously, we restrict the current scope to a single plan to maximize the computational efficiency of the parallel physics simulation.

\subsection{Symbolic Task Planning}

The objective of the symbolic task planning step is to provide a candidate symbolic plan $\pi$ that is 1) general enough to allow many possible realizations, and 2) informative to the subsequent plan realization step. 
To this end, we deliberately model the set of possible actions as general as possible.
For example, in \Cref{fig:intro}, we consider actions such as  "move", "pick", or "place", without specific references to the environment, e.g., to objects or regions. Thus, without requiring expert knowledge of the environment.

We use PDDL \cite{mcdermott_pddl-planning_1998} to model the symbolic planning problem.
This choice gives us an action-focused notation with flexibility in the level of detail considered in the abstractions. 
However, we need to add additional information to distinguish between different parameters for the actions. 
More formally, we define the task planing problem with:

\paragraph{Parameters}

We define the set of symbolic parameters $\mathcal{Z} = \{ z_1, z_2, \ldots \}$. 
Each parameter $z_j \in \mathcal{Z}$ is defined as a tuple $z_j = \langle \text{sym}_j, \mathcal{M}_j, \mathcal{C}_j \rangle$, where: 
$\text{sym}_j$ is the unique identifier (e.g., $\textit{Table1}, \textit{Cube}, \textit{Start}$) used in the PDDL domain. 
$\mathcal{M}_j$ is the underlying manifold (e.g., $\mathbb{R}^2, SE(3)$) defining the continuous space of the parameter. We use $\symb$ to denote a parameter that requires no realization, indicating it is used solely for symbolic logic and not as an optimization variable.
$\mathcal{C}_j$ is a set of constraint functions $h(z) \leq 0$ defining the valid range or region (e.g., circular region around a table).
During symbolic planning, $z_j$ is treated as a discrete token. During plan realization, we optimize over values in $\mathcal{M}_j$ only for parameters where $\mathcal{M}_j \neq \symb$

\paragraph{Action Schema}
The robot can execute different controllers $\controller_i$.
The parameters of a controller result in different actions, e.g., the different $\textit{moveTo}$ actions in Figure~\ref{fig:realization_diagram}a.
We define action schemas as abstractions of high-level skills (e.g., grasping or placing) and the associated controllers and parameters. 
Action schemas $\alpha_i$ are defined by the tuple 
\begin{equation}
\alpha_i = (\mathcal{Z}_{\alpha_i},\; \controller_{\alpha_i},\; J_{\alpha_i}(\mathbf{x}_{t_{0,i}:T_i}),\; \text{success}_{\alpha_i}, \psi_{\text{pre}}, \psi_{\text{eff}}),
\label{eq:action}
\end{equation}
where $\mathcal{Z}_{\alpha_i} \subseteq \mathcal{Z}$ is the subset of parameters consistent with the action's type of input.
For example, if the action schema uses a point in $\mathbb{R}^2$ for its tracking controller, the type of parameter must also be in $\mathbb{R}^2$.

The $\controller_{\alpha_i}$ uses the parameters to calculate the action motion and control the robot. The action schema has a cost $J_{\alpha_i}(\mathbf{x}_t)$, which can be continuously evaluated or only once when the action is finished.
We can determine an action's success by evaluating the Boolean function $\text{success}_{\alpha_i} \in \{ \textit{True},\textit{ False}\}$, depending on the executed motion.

Following the PDDL convention, $\psi_{\text{pre}}$ is the set of precondition predicates, which are evaluated on the symbolic parameters. The action can be executed if the preconditions are met. Similarly $\psi_{\text{eff}}$ is the set of effect predicates that abstract the effects of a controller in the environment by describing which predicates change value after execution. By chaining actions of compatible effects and preconditions, we can limit the possible sequences of actions to the ones that are at least feasible at the abstract level. 

As the plan's feasibility is only verified in the realization step, see \Cref{sec:optimization_of_the_actions}, we do not require the $\psi_{\text{pre}}$ and $\psi_{\text{eff}}$ predicates to accurately model complex physical interactions in the environment. 
For example, when performing a grasping action, we do not enforce that a grasp is physically stable. 
Instead we limit the predicates to model logic at the symbolic level, such as not being able to perform the grasp action if the robot gripper is not empty. 

When a compatible assignment of parameters $\mathcal{Z}_{a_i} \subseteq \mathcal{Z}$ is made to instantiate an action schema, we obtain a symbolic action $a_i$.
As an example, consider the symbolic actions resulting from a different choice of parameters for the schema: $\textit{moveTo}(\cdot,\textit{Exit}) ,\; \textit{moveTo}(\cdot,\textit{Table}_1)$. 
The action is \emph{realized} when a value is chosen for its parameters $\mathcal{Z}_{a_i}$.

A symbolic plan is a sequence of symbolic actions $\pi = ( a_1, \ldots, a_k )$ with its associated parameters, $\mathcal{Z}_\pi = ( \mathcal{Z}_{a_1}, \ldots, \mathcal{Z}_{a_k} )$. 
We use a standard symbolic planner \cite{helmert_fast_2006} to generate the symbolic plan that accomplishes goal predicates $\tilde{\psi}_g$, which we assume models  $\psi_g(\mathbf{x}_T)$ at the symbolic level.

\subsection{Plan Realization}
\label{sec:optimization_of_the_actions}

Once the solver generates a candidate symbolic plan, we need to find a feasible and low-cost realization of the actions to execute the plan on the real robot. 
A plan is realized when all its actions are realized, and it is feasible if all actions can be executed and the goal specification  $\psi_g(\mathbf{x}_T)$ is achieved. 
The total cost of a realized plan is the sum of all realized action costs plus a final cost. 
\begin{problem}[Symbolic Plan Realization]
\label{prob:realization}
   Given a symbolic plan  $\pi = a_{1:k}$, goal specification $\psi_g(\mathbf{x}_T)$ and system dynamics $f$, find a realization of $Z_\pi = Z_{a_{1:k}}$ that solves
   \begin{equation*}
\begin{aligned}
\min_{\mathcal{Z}_{a_{1:k}}} \quad & J_{\text{end}}(x_{T} , \mathcal{Z}_{a_{1:k}} ) + \sum_{i=1}^k \sum_{t=t_{0,i}}^{T_i} J_{a_i}(x_t)\\
\textrm{s.t.} \quad & \text{success}_{a_i} = \textit{True}, \; \forall \; a_i \in \pi,   \\
  & \psi_g(\mathbf{x}_T) = \textit{True},   \\
  &  x_{t+1} = f(x_t, a_i),   \\
\end{aligned}
\end{equation*}
where $t_{0,i}$ and $T_i$ represent the start and end time of $a_i$. 
\end{problem}

In order to optimize the values of the parameters $\mathcal{Z}_\pi$ of a given plan $\pi$ in \Cref{prob:realization}, we use Cross-Entropy optimization (CE) \cite{botev_chapter_2013}. The method iteratively refines the estimation of the optimal parameter values by using importance sampling. 

In CE optimization, we assume that there exists a target distribution over the parameter space from which samples have a high chance of being low cost. 
Our aim is to move an approximated distribution closer to this target distribution after each iteration, to generate low cost parameter samples. 
To this end, we assign a probability distribution to each parameter of an action $\mathbf{z}_{k} \sim \phi(\mathbf{\theta}_{k}) \;\forall \; \mathbf{z}_k \in Z_{a_i}$ parametrized by $\theta_k$. Let $\phi(\theta_\pi)$ be the resulting stacked vector of probability distributions of all the parameters of a plan $\pi$. For convenience, we refer to sampling all the parameters of the plan as sampling the vector $Z_\pi \sim \phi(\theta_\pi)$. 

After sampling an $n_{\text{env}}$-sized batch of parameter values $\mathcal{Z}_\pi$, we roll out the plans in parallel using a physics simulator, and subsequently evaluate the samples feasibility and performance. The top $N_e$ elite samples, that is the feasible ones with lowest cost, are used to find a new $\theta_\pi$ such that the KL divergence with respect to the target distribution is minimized~\cite{botev_chapter_2013}. With the updated approximated distribution, we can sample again and repeat until convergence. This approach allows for gradient free optimization and to consider the discontinuities and nonlinearity of the dynamics.

\Cref{alg:ce} outlines our CE optimization algorithm. 
The initial sampling stage is critical for the algorithm's performance, as ideally it provides enough feasible samples representing the potential multi-modal realizations. 
Consider, for example avoiding an obstacle from the left, or from the right; these are two realizations that may exist in disconnected regions of the parameter spaces. For this reason, the initial distribution $\phi(\theta^0_\pi)$ used in \Cref{alg:ce:initial} is uniformly distributed inside the region of the parameter $h(z_k) \leq0$, which acts as an initial guess of areas where solutions may exists. 
\Cref{fig:realization_diagram}a illustrates examples of such parameter regions, e.g., a circular approach region around the table. 
In subsequent iterations of the algorithm, the constraints on parameter values may not be enforced, as we assume that the next samples will not deviate excessively from the previous feasible samples. 
In line \ref{alg:CE:update_theta}, if the number of feasible solutions $n_{\text{feasible}}$ is lower than $N_e$, only the $n_{\text{feasible}}$ samples are used to update the distribution. If only one solution is found, the successful sample is used with added artificial noise to enforce exploration around the sample in the subsequent iterations. 

In cases where an action's parameter space is high-dimensional and its region is poorly chosen, such as grasp poses, we may not find  feasible plan realizations during the initial sampling step. Even if a valid grasping pose is sampled, it may not satisfy $\psi_g(\mathbf{x}_T)$ in combination with the other realized plan parameters.  
In these cases, we use the plans where at least all actions are feasible, i.e., $success_{a_i} = \text{True}, \; \forall a_i \in \pi$, for the update of the sampling distribution. 
We hypothesize that the parameters of a plan where all actions are individually successful are strongly correlated to parameters that render the whole plan feasible, i.e., $\psi_g(\mathbf{x}_T) = \text{True}$.
As a result, subsequent sampling iterations have a higher chance of generating feasible plan realizations.

\begin{algorithm}[t]
\caption{Plan Parameter Cross-Entropy Optimization}
\begin{algorithmic}[1]
\Statex \textbf{Inputs:}
\State $\pi \text{: Symbolic plan}$
\State $\mathcal{Z}_\pi \text{: Symbolic parameters used by the actions}$
\State $\phi(\theta^0_\pi)\text{: Parameterized initial distribution}$ \label{alg:ce:initial}
\State $n_{env}(j) \text{: Number of parallel plans for each iteration } j$
\State $N_e \text{: Number of important samples}$
\Statex \textbf{Main loop:}
\State $j = 0$ \Comment{CE iteration counter}
\While{$\theta^j_\pi$ not converged}
\State $S \gets \text{Sample } n_{env}(j) \text{ times } Z_\pi \sim \phi(\theta^j_\pi)$
\State $C \gets$ Evaluate successful plan costs of samples in $S$
\State $\theta^{j+1}_\pi \gets$ Update distribution from top $N_e$ samples \label{alg:CE:update_theta}
\State $j = j+1$
\EndWhile
\State \Return Best sample $Z_\pi \in S$, and final $\theta^{j+1}_\pi$
\end{algorithmic}
\label{alg:ce}
\end{algorithm}

\section{Results}

\begin{figure*}[t!]
\centering
    \includegraphics[width=.85\linewidth]{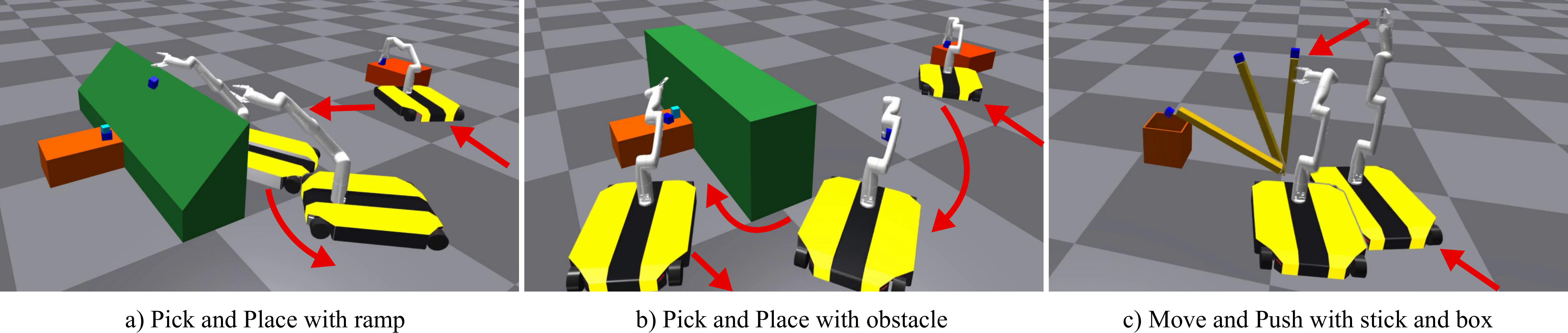}
    \caption{Simulation results of: a) plans for ramp-shaped obstacle, and b) without ramp. When the ramp is present, the robot uses it to let the cube slide, avoiding going around the obstacle. When the obstacle is a box, the robot only finds solutions that go around it. In c), the move and push scenario.}
    \label{fig:sim_solutions}
\end{figure*}

We evaluate our method for a pick and place task of a cube with a mobile manipulator, and for a move and push task. 
These scenarios were selected to demonstrate the method's capability to resolve complex hybrid dynamics
where defining explicit analytical constraints is non-trivial.
We show simulation and real-world results, \Cref{fig:sim_solutions} and \Cref{fig:real_experiments}.

\subsection{Experimental setup}

\paragraph{Robot and dynamics model}
The robot in our experiments is a \href{https://clearpathrobotics.com/dingo-indoor-mobile-robot/}{Dingo-O with Kinova 6DoF arm and gripper}.
To model the dynamics and complex contacts of the robot interacting with objects in the environment, we use the GPU-based physics simulator IsaacGym \cite{makoviychuk2021isaac}. The simulator allows for computationally efficient parallelized execution of plans.

To keep computation efficient, we require the use of robot controllers that can be evaluated efficiently and in parallel, together with the simulator. For our experiment, we consider: 

\paragraph{Whole-body control} We employ Geometric fabrics for fast computation of motion for the base and the robot arm when end-effector pose control with collision avoidance is required. 
Geometric fabrics \cite{ratliff_optimization_2020, spahn_dynamic_2023} are based on Riemannian Motion Policies (RMPs) \cite{ratliff2018riemannian} where the desired behaviors of the robot, such as collision avoidance and convergence to a goal pose, are composed into a differential equation:
\begin{equation}
\tilde{\boldsymbol{M}}(\boldsymbol{q}, \dot{\boldsymbol{q}}) \ddot{\boldsymbol{q}}+\tilde{\boldsymbol{f}}(\boldsymbol{q}, \dot{\boldsymbol{q}})+\gamma \partial_{\boldsymbol{q}} \psi+\boldsymbol{B} \dot{\boldsymbol{q}}=0 ,
\label{eq:fabrics}
\end{equation}
where $\boldsymbol{M}$ and $\boldsymbol{f}$ encode the system that is forced by a potential $\psi$ towards a goal with attractor weight $\gamma$ and damping $\boldsymbol{B}$.
Solving \eqref{eq:fabrics} yields the trajectory generation policy $\ddot{\boldsymbol{q}}=\tilde{\pi}(\boldsymbol{q}, \dot{\boldsymbol{q}})$ for whole-body control of the robot, which allows for parallelized evaluation. 

\paragraph{Path planning for the base} We employ \textit{A*} \cite{Hart1968} to compute collision-free paths for the mobile base when moving between locations. The generated waypoints are tracked with a PID controller and evaluated in parallel. 

\paragraph{TAMP problem}

Our pick and place task with one object and two tables is illustrated in \Cref{fig:realization_diagram}a.
The goal specification is for the cube to be on Table 2, and the robot needs to be in the exit zone by the end of the plan. Between the tables, there is an obstacle, the geometry of which could be exploited to find better solutions. 
We consider two setups: Setup 1) the obstacle has a slanted side, effectively creating a ramp, and  setup 2), the obstacle has no slanted side, making it a  rectangular prism, which we call box. \Cref{fig:realization_diagram}a  shows a diagram of the environment setup with the ramp obstacle.

We define the following symbolic parameters that represent regions and grasps of interest

\begin{itemize}
    \item  $Table_1 \in \mathbb{R}^2$. Represents the region of points around Table 1. The initial uniform sampling is taken within a ring $(r= 1.1,\; R=1.5)$m centered around the table.
    \item  $Table_2 \in \mathbb{R}^2$. Region of points around Table 2. The initial uniform sampling is taken from within a ring $(r=1.2 ,\; R=1.7)$m centered around the table.
    \item $Cube \in \text{SO}(3)$. Represents the orientation of a grasp pose, located in the center of the object to be picked. The orientation is encoded as a quaternion initially sampled uniformly within a 0.3 size hyper-rectangle centered around a top-down grasp pose.
    
    \item $Target \in \text{SE}(3)$. Represents the position and orientation of the end effector when the grasped object is released. It is encoded by a location $p\in \mathbb{R}^3$, initially sampled from a hyper-rectangle of size $(0.2,0.4,0.6)$m, located on top of the table and partially on the obstacle, and a quaternion initially sampled as in $Cube$.
    \item $Exit \in \mathbb{R}^2$. Represents a rectangular area of size (1,1)m where the robot needs to be at the end of the plan. The values are initially sampled uniformly in the rectangle. 
\end{itemize}

We define the following action schemata to abstract moving, picking and placing and generate the possible symbolic actions together with the previously defined parameters. 

\begin{align}
     & move\_to(\{ p_1 \in \symb ,\; p_2 \in \mathbb{R}^2 \}, \text{A*}, J_m(x_t), succ_m(p), \label{eq:moveto} \\
     & \phantom{move\_to( } \psi_{\text{pre}} = \{ robot\_at(p_1) \}, \notag \\
     & \phantom{move\_to( } \psi_{\text{eff}} = \{ robot\_at(p_2), \; \neg \; robot\_at(p_1) \}) \notag 
\end{align}
\vspace*{-0.70cm}
\begin{align}
     & grasp(\{ q \in SO(3), p \in \symb \}, \text{Fabrics}, J_g(x_t), succ_g(q),  \label{eq:grasp} \\
     & \phantom{grasp} \psi_{\text{pre}} = \{ robot\_at(p), \; object\_at(q,p), \; ee\_empty  \}, \notag \\
     & \phantom{grasp} \psi_{\text{eff}} = \{ \neg \; object\_at(q,p), \neg \; ee\_empty \}) \notag 
\end{align}
\vspace*{-0.70cm}
\begin{align}
     & place(\{ q \in SE(3), \; p \in \symb \}, \text{Fabrics}, J_p(x_t), succ_p(p),  \label{eq:place} \\
     & \phantom{place( } \psi_{\text{pre}} = \{ robot\_at(p), \neg \; ee\_empty  \}, \notag \\
     & \phantom{place( } \psi_{\text{eff}} = \{ object\_at(q,p), \; ee\_empty \}) \notag
\end{align}
The robot arm remains static during $moveTo$. The cost $J_m$ is set to $1$ for every time step the action is not successful. The success condition is true if the mobile base is at the goal location within a tolerance $succ_m(p_2): \norm{x_{base} - p_2} \leq \epsilon $ 

For \eqref{eq:grasp} and \eqref{eq:place}, we use as input parameters an orientation and a grasp pose respectively. Both use the full body (mobile base + arm) parallelized geometric fabrics controller. As before, the cost is simply $1$ every time step the action is not finished. The success condition is true if the end effector has reached the grasp/place pose within a tolerance $succ_g(q): \norm{q - p_{ee}} \leq \epsilon $. For grasping, we additionally check for the gripper to enclose the object. 

To generate candidate symbolic plans, we consider the symbolic goal $robot\_at(Exit)\;AND \; object\_at(Table_2) $. The resulting realized plan is considered successful if $\psi_g(\mathbf{x}_T): x_{robot}(T) \in Exit$ and $x_{cube} \in \mathcal{X}_{Tabe\_Surface}$, i.e. the cube is on the upper surface of Table 2, and the robot is at the exit region.
Notice how the requirement of a successful plan does not enforce the robot to be in the annular regions $Table_1, Table_2$ after executing the move to actions. The regions serve as an initial guess for sampling. 

We create a candidate symbolic plan using the off-the-shelf PDDL solver \cite{Helmert_2006}. The resulting plan accomplishes at the symbolic level the goal of moving the cube to the second table, and ends with the robot in the designated exit area. 

The plan sequence is defined as follows, where parameters in \textbf{bold} denote continuous variables optimized by our framework, while non-bold parameters represent fixed symbolic context ($\mathcal{M} = \symb$):
\begin{multline}
     \pi_{p\&p} = \mathrm{moveTo}(Start, \mathbf{Table_1}), \mathrm{grasp}(\mathbf{Cube}, Table_1), \\
     \mathrm{moveTo}(Table_1, \mathbf{Table_2}),\; \mathrm{place}(\mathbf{Target}, Table_2), \\
     \mathrm{moveTo}(Table_2, \mathbf{Exit})
\end{multline}

Similarly, for the move and push problem, we have an open box, a wooden rod, and a red block on its top. 
The goal is to place the red block inside the box as fast as possible. The candidate symbolic plan is 
\begin{equation}
\pi_{m\&p} = \mathrm{moveTo}(Start,  \mathbf{rod_{loc}}), \mathrm{push}(rod_{loc} , \mathbf{rod_{p}}) 
\end{equation}

We initialize the symbolic location $rod_{loc}$ as an annular region around the rod.  
The \textit{push} action is parameterized by a target end-effector pose $rod_{p} \in SE(3)$. Its controller reaches the target and opens the gripper. The symbolic preconditions require the robot to be located at $rod_{loc}$ and the block to be on the rod.
This pose, $rod_{p}$, is initially sampled inside a cuboid centered around the rod, as seen in \Cref{fig:real_experiments}


\paragraph{Algorithm setting}
We run the CE \Cref{alg:ce}, by initially sampling values uniformly within the bounds of the parameters. For subsequent iterations, we model the distribution of the parameters as an independent normal distribution. The resulting $\mu$ and $\sigma$ of the elite parameter samples are used for the next iteration. We initially sample $n_{envs} = 3000$ and decrease linearly to $300$ in iteration 10, constant afterwards. $N_e = 50$ elite samples are used.
We repeat each experiment for 10 different random seeds.

\begin{figure*}[t]
    \centering
    \begin{minipage}[t]{0.30\textwidth}
        \centering
        \includegraphics[width=\linewidth]{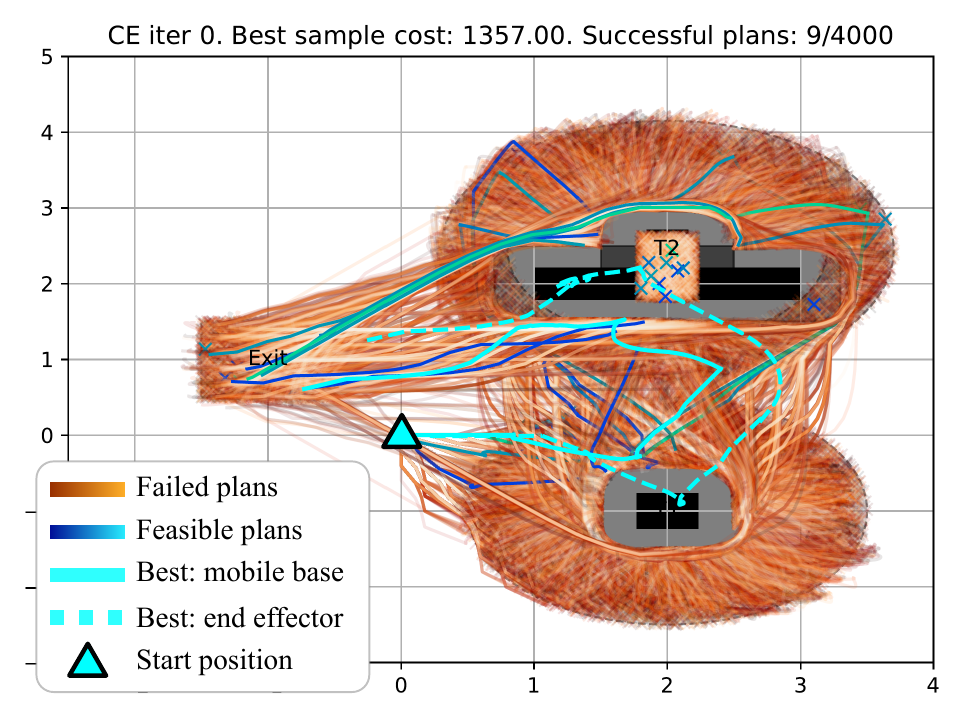}
        {\footnotesize \text{(a)} Iteration 0}
    \end{minipage}%
    \hfill
    \begin{minipage}[t]{0.30\textwidth}
        \centering
        \includegraphics[width=\linewidth]{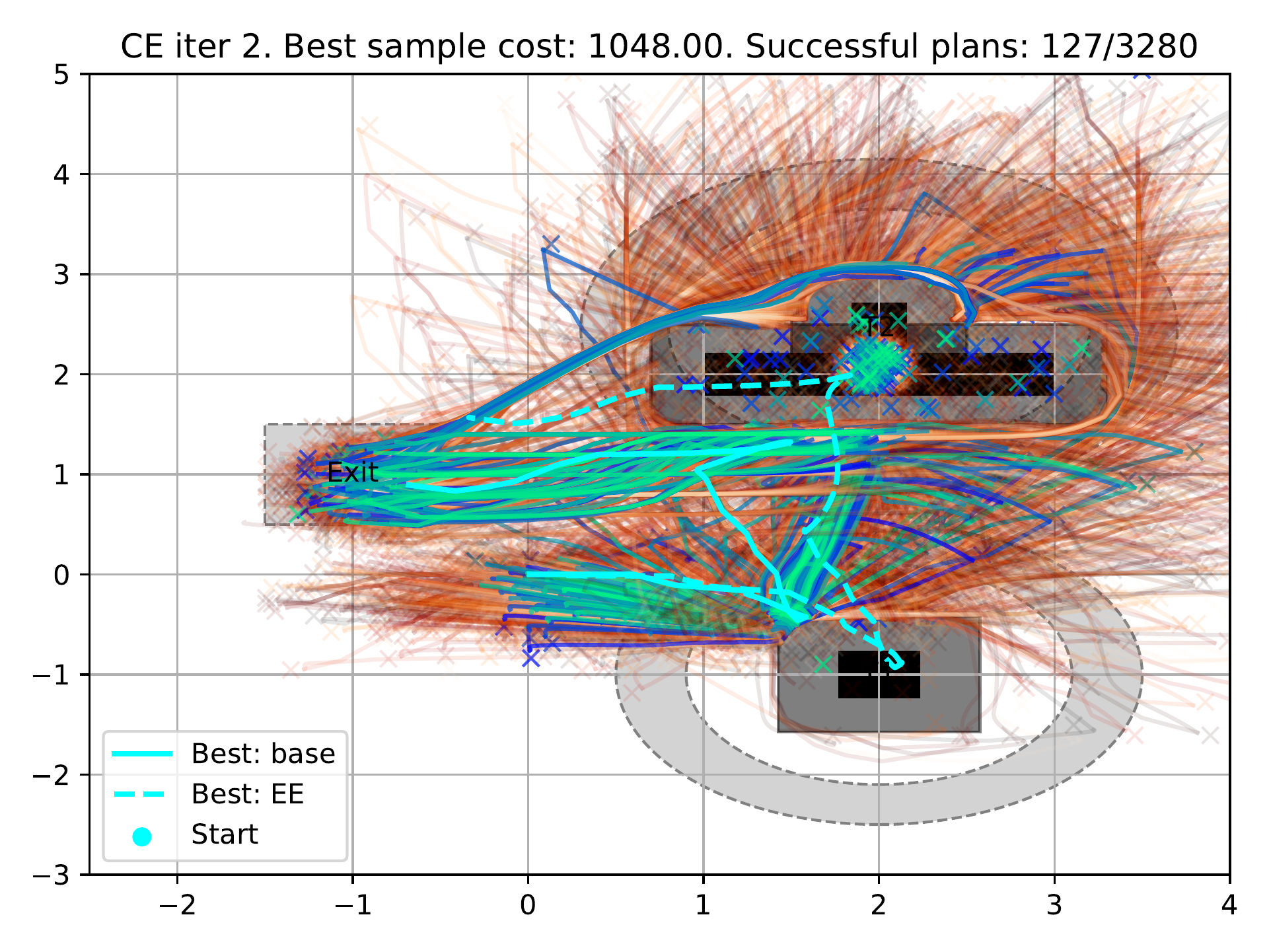}
        {\footnotesize \text{(b)} Iteration 2}
    \end{minipage}%
    \hfill
    \begin{minipage}[t]{0.30\textwidth}
        \centering
        \includegraphics[width=\linewidth]{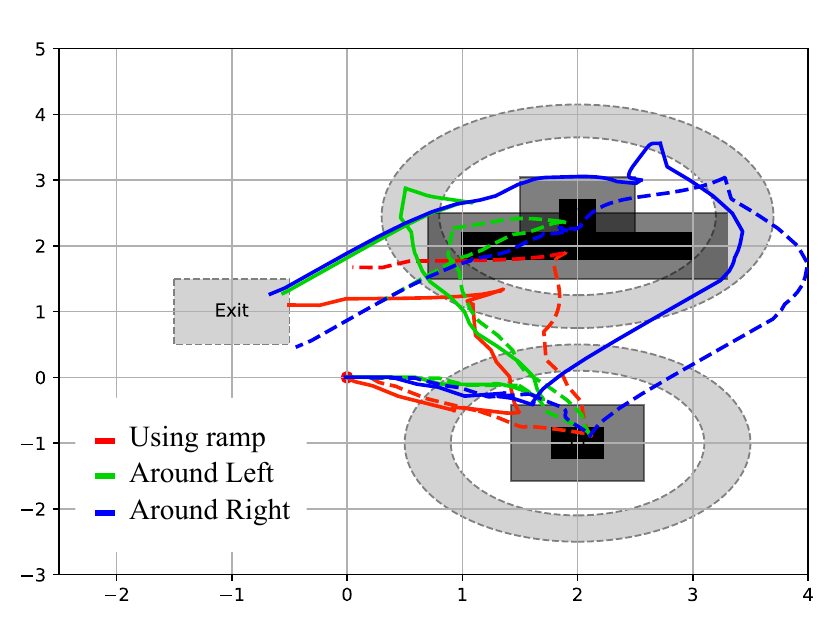}
        {\footnotesize \text{(c)} Final iterations for different runs.}
    \end{minipage}
    
    \caption{Plans simulated from the sampled parameters for different iterations of the CE optimization. The trajectories displayed in orange tones show the trajectory of the mobile base of failed plans. The trajectories in cold tones show the successful plans. The best-sampled plan is shown in cyan, both base and end effector trajectory (EE), in continuous and dashed lines, respectively. The cross markers show the value of the parameter. The parameter areas are in clear grey. The tables and obstacles are in black, with 20cm of additional padding, as used by the A* algorithm.}
    \label{fig:samples}
\end{figure*}

\subsection{Simulation Results}

\Cref{fig:samples} shows sampled plan realizations at different CE stages for the pick and place problem (ramp setup). The sampled parameters and trajectories are projected onto the ground plane. 
The best sample's end-effector trajectory is overlaid.

In the pick and place problem with the ramp obstacle, the best solution exploits the geometry by sliding the cube down the slope to the table, avoiding going around it (\Cref{fig:sim_solutions}a).
When the obstacle is a box instead, the robot goes around it to directly place the cube (\Cref{fig:sim_solutions}b).

\begin{figure*}[t]
    \centering
    \begin{minipage}[t]{0.30\textwidth}
        \centering
        \includegraphics[width=\linewidth]{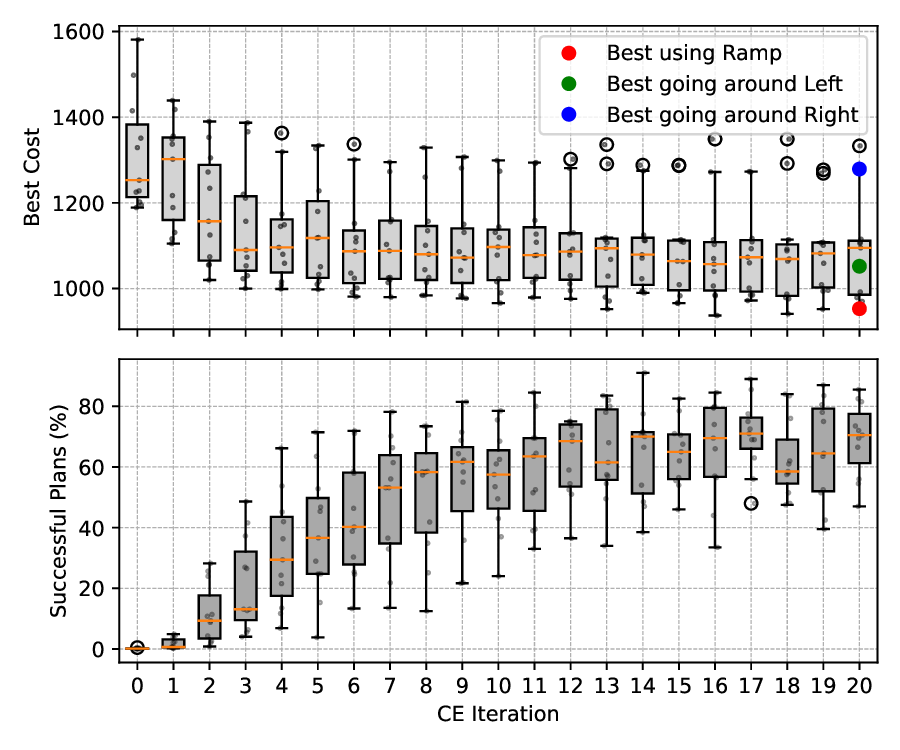}
        {\footnotesize \text{(a)} Pick and place with ramp obstacle}
    \end{minipage}%
    \hfill
    \begin{minipage}[t]{0.30\textwidth}
        \centering
        \includegraphics[width=\linewidth]{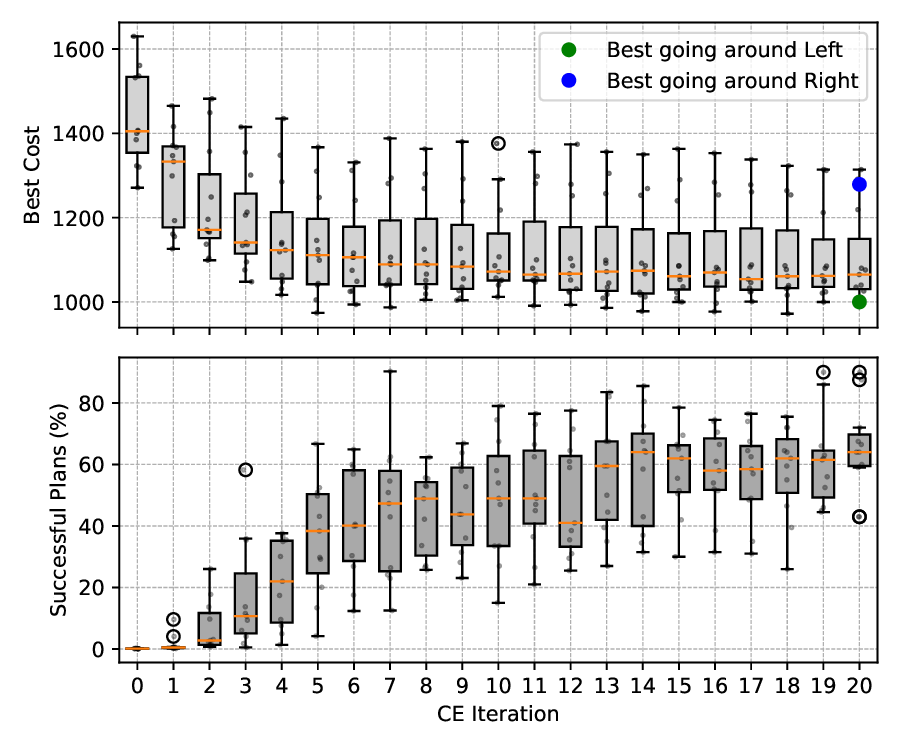}
        {\footnotesize \text{(b)} Pick and place with box obstacle}
    \end{minipage}%
    \hfill
    \begin{minipage}[t]{0.30\textwidth}
        \centering
        \includegraphics[width=\linewidth]{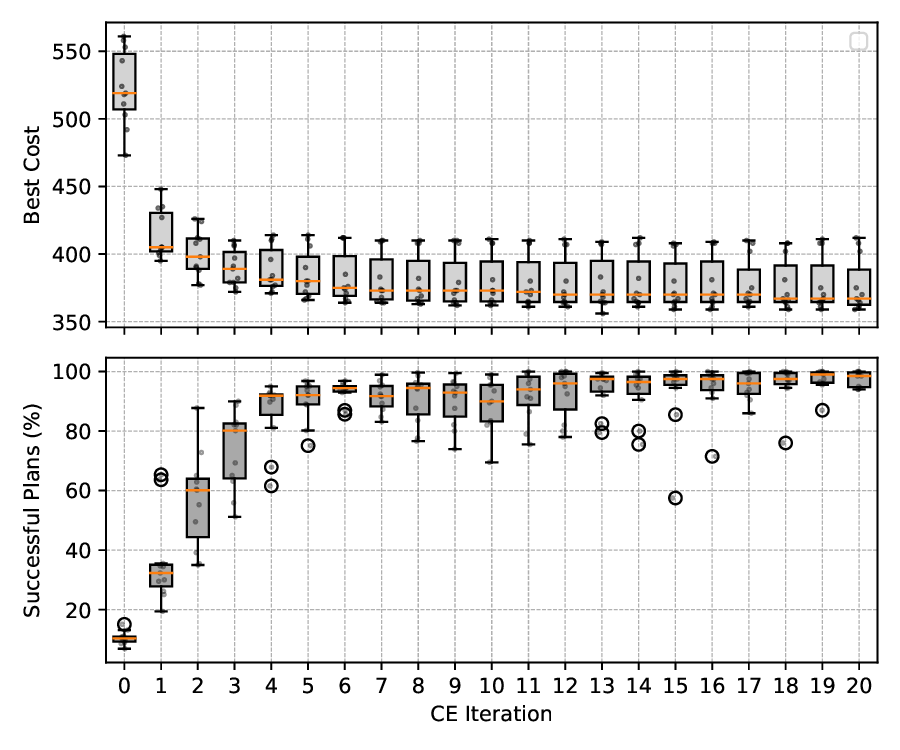}
        {\footnotesize \text{(c)} Move and Push problem.}
    \end{minipage}
    \caption{Top: Cost evolution of the best-sampled plan realization for each scenario. Bottom: Percentage change of sampled, feasible plans.}
    \label{fig:cost_evolution}
\end{figure*}

\Cref{fig:cost_evolution}a-b shows the success rate and cost of the best sampled plan over CE iterations for both setups.
We observe that initially, only $\approx 0.1\%$ of the sampled plans are feasible. 
Feasible values for parameters like grasp and place poses occupy highly constrained, disconnected regions of the parameter space, causing low initial success rates.
On the other hand, parameters such as locations are more often successful, as the controller used can accomplish the move action reliably for most parameter's values. 
Over iterations, successful plan realization samples significantly increase and costs decrease.
For these reported metrics, we utilized a fixed budget of 20 iterations to ensure asymptotic stability, though we observe that the solution cost typically converges ($<1\%$ variance) within the first 5-10 iterations.

We also observe the multi-modality of the realizations found in the pick and place problem. The solution modes can be divided in three, ordered by increasing average cost: 1) using the ramp (only possible with the ramp), 2) going around the obstacle from the left, and 3) going around the obstacle from the right. \Cref{fig:samples}c shows a converged example for each mode.
As we use normal distributions for sampling the parameters, they eventually collapse to single local minima corresponding to a mode. If the initial samples cover the full range of modes equally, thus providing information about the different costs, \Cref{alg:ce} converges to the mode that provides the best costs. However, as few feasible plans are found in the initial iterations, some modes may be overrepresented, and the updated distribution will move towards them even if they create higher cost realizations. In \Cref{fig:cost_evolution}a, with the ramp, we observe that out of the 10 final solutions, 6 use the ramp, 3 go around the left, and 1 goes around the right. For the box setup, \Cref{fig:cost_evolution}b, we observe 7 solutions going around the left, 3 going around the right. 

Evaluating the computational cost on an NVIDIA 4060 (Laptop), we find the bottleneck is parallel controller execution ($\approx 80\%$ of total time) compared to physics simulation ($\approx 19\%$).
However, the time per iteration decreases drastically as the optimization converges. For the pick and place scenario, the compute time drops from $\approx 6$ min (Iteration 0) to $50$s (Iteration 10+) as the number of active environments decreases and trajectories become more efficient. Similarly, the Push scenario decreases from $1.5$ min to $15$s per iteration.
The cost of the plan realizations follows a steep decrease during the initial iterations \Cref{fig:cost_evolution}, which allows the possibility to obtain a low cost solution with relatively few iterations.

\subsection{Real World Experiments}

\begin{figure}[t]
    \centering
    \includegraphics[width=0.85\linewidth]{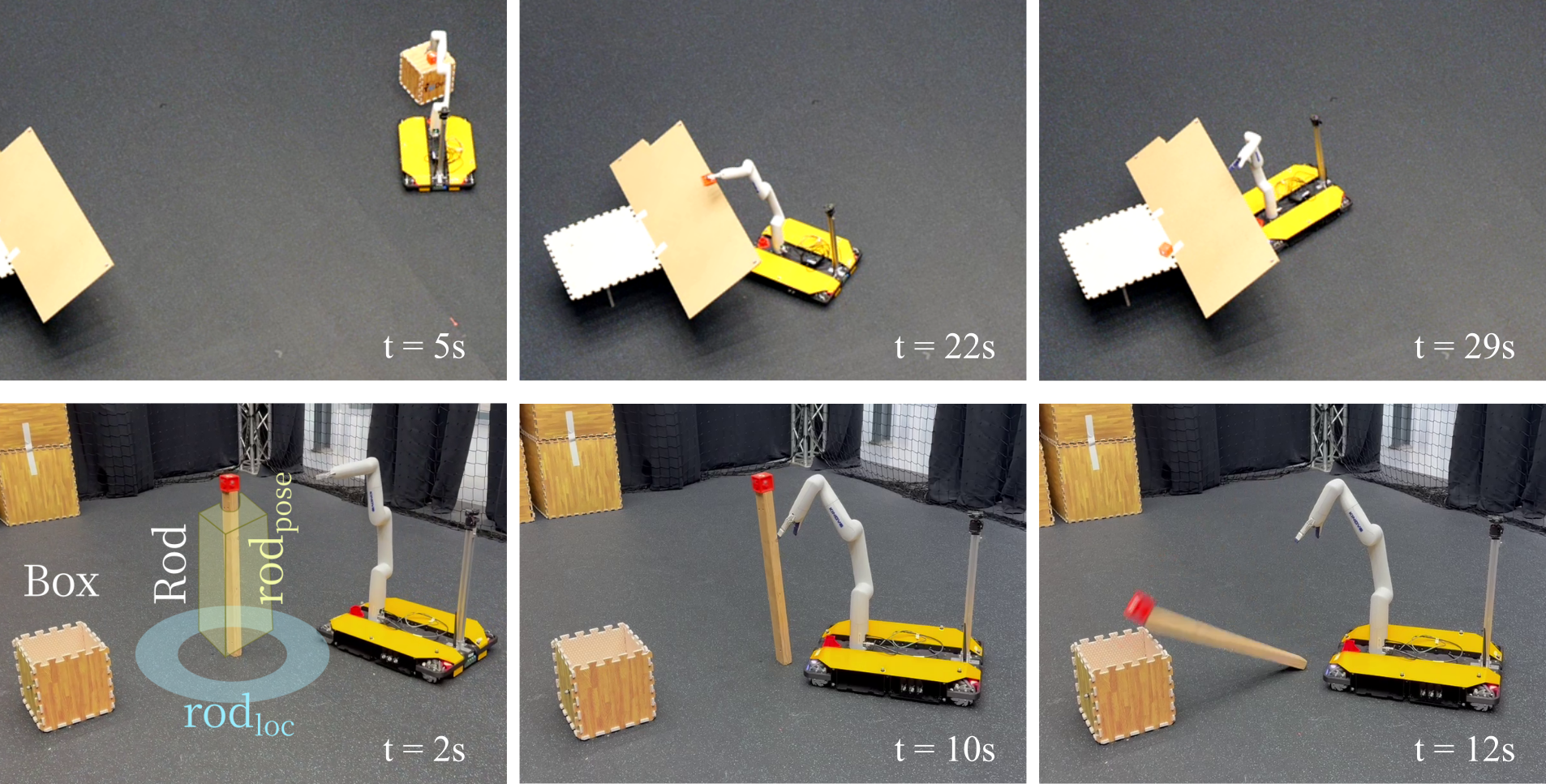}
    \caption{Plan executed in the real system, using the same low level controllers. (Top row): Pick and place scenario with the ramp-shaped obstacle. (Bottom row): Move and Push problem.}
    \label{fig:real_experiments}
\end{figure}

We report results for 1) the pick and place (ramp setup), and 2) move and push problem on a real-world system. 
To bridge the sim-to-real gap, we rely on three design choices. First, identical closed-loop controllers in simulation and reality compensate for minor dynamic mismatches. Second, object-centric reference frames to ensure actions remain valid relative to the object even if global localization varies. Third, we utilize Domain Randomization (DR) to robustify plans against bounded uncertainty in friction or mass \cite{pezzato2025samplingbasedmodelpredictivecontrol}, though deterministic optimization proved sufficient given accurate motion-capture estimation.
Simulator inputs include geometries of the objects in the environment and their locations. \Cref{fig:real_experiments} shows key execution moments.

The real-world execution consistently matched the optimal simulation strategy (\Cref{fig:sim_solutions}). While the pick-and-place task took longer (35 s vs. 21 s) due to ground friction and less capable actuators than in simulation, the move-and-push task showed only a 3 s difference. 
The solution showcases that complex contacts and dynamics such as sliding a cube can be considered at the plan realization stage in a real system, without injecting expert knowledge at the symbolic stage.

\section{Conclusions} 

\label{sec:conclusion}

Our work computes plan realizations of task plans while implicitly
accounting for robot dynamics and contacts with the
environment. 
We employ a parallelized physics simulator and cross-entropy optimization to sample the
parameters used by the controllers, or actions, and obtain feasible and low-cost solutions. 
The results of our simulations and real-world experiments showcase that our method is capable of finding low-cost plan realizations within a few CE iterations and with little prior information. 
The generated solutions consider geometrical constraints, such as the inability of the robot to reach objects from certain positions. 
Instead of explicitly modeling these constraints in the problem, our method implicitly models them as a consequence of the physics simulations. 
This allows us to consider the effects of the utilized controllers, e.g., as seen when our mobile base is repositioned while pushing against the table during grasping and placing. 
This behavior resulting from environment contacts and robot controllers could not be modeled in previous sampling-based methods \cite{garrett_pddlstream_2020,thomason_task_2022}.

\textbf{Limitations:} Despite GPU parallelization, evaluating full-horizon physics rollouts remains computationally expensive, currently limiting real-time planning, or evaluating different symbolic plans efficiently. Furthermore, the approach relies on informative initial regions; without these, the optimizer may fail to converge in sparse solution spaces.

\bibliographystyle{IEEEtran}
\bibliography{IEEEabrv, references_pruned}

\end{document}